# Multistage SFM: A Coarse-to-Fine Approach for 3D Reconstruction

Rajvi Shah[a], Aditya Deshpande[a,b], P J Narayanan[a]

[a]*Center for Visual Information Technology, IIIT Hyderabad, India*
[b]*Department of Computer Science, University of Illinois Urbana-Champaign, Illinois, USA*

**Abstract**

Several methods have been proposed for large-scale 3D reconstruction from large, unorganized image collections. A large reconstruction problem is typically divided into multiple components which are reconstructed independently using structure from motion (SFM) and later merged together. Incremental SFM methods are most popular for the basic structure recovery of a single component. They are robust and effective but strictly sequential in nature. We present a multistage approach for SFM reconstruction of a single component that breaks the sequential nature of the incremental SFM methods. Our approach begins with quickly building a coarse 3D model using only a fraction of features from given images. The coarse model is then enriched by localizing remaining images and matching and triangulating remaining features in subsequent stages. The geometric information available in form of the coarse model allows us to make these stages effective, efficient, and highly parallel. We show that our method produces similar quality models as compared to standard SFM methods while being notably fast and parallel.

## 1. Introduction

The field of large-scale structure from motion (SFM) and 3D reconstruction has seen a steady progress in the past decade. Snavely et al. (2006) presented Photo Tourism, a system for navigation, visualization, and annotation of unordered Internet photo collections based on a robust method for incremental structure from motion (Brown and Lowe, 2005). Snavely's incremental SFM software, Bundler, is widely used since then. Bundler is a robust and effective system but one with quadratic and cubic costs associated with exhaustive pairwise feature matching and bundle adjustment. The effectiveness of this system however inspired attempts to yield city-scale 3D reconstructions in under a day by identifying many independent sub-tasks involved, and leveraging multi-core clusters and GPUs to parallelize these tasks (Agarwal et al., 2009; Frahm et al., 2010). Since then, researchers have continued to improve the large-scale reconstruction pipeline in many ways.

The large-scale SFM pipeline can broadly be divided into five steps (see section 2 and Table 1). In steps 1 and 2, a large reconstruction problem is broken down into multiple components based on the image connections. Steps 3 and 4 involve pairwise feature matching and 3D reconstruction of a single component; these were at the core of PhotoTourism. While the original incremental SFM method is still widely used, hierarchical and global methods that differ significantly have been proposed subsequently. In this paper, we present a *multistage* method for steps 3 and 4 that provides greater efficiency and completeness to SFM. Our method builds on several prior efforts for reconstructing a single component typically of about a 1000 pictures, representing a facade, a building, a street, or parts of it.

The motivation behind our multistage approach is akin to the coarse-to-fine strategies of several vision algorithms. We wish to quickly recover a coarse yet global model of the scene using fewer features and leverage the constraints provided by the coarse model for faster and better recovery of the finer model using all features in the subsequent stages. Feature selection for recovering the coarse model can be based on several aspects. We use the scales of SIFT features for this. The coarse model provides significant geometric information about the scene structure and the cameras, in form of point-camera visibility relations, epipolar constraints, angles between cameras, etc. By postponing the bulk of the processing until after the coarse model reconstruction, our approach can leverage rich geometric constraints in the later stages for effective, efficient, and highly parallel operations. We demonstrate that the proposed staging results in richer and faster reconstructions by using more compute power on several datasets.

Summarily, we make the following contributions: (i) we propose a coarse-to-fine, multistage approach for SFM which reduces the sequentiality of the incremental SfM pipeline; (ii) we demonstrate applicability of fast 3D-2D matching based localization techniques in context of SFM and utilize it for simultaneous camera pose estimation; (iii) we present an intelligent image matching strategy that utilizes the point-camera visibility relations for filtering image pairs to match and we also propose a fast and effective algorithm for geometry-aware feature matching that leads to denser correspondences and point clouds.

| Step | Category | | Reference |
|---|---|---|---|
| **Step 1**<br>Select Image<br>Pairs to Match | 🟥 Image Retrieval based<br>🟥 Feature Matching based<br>🟪 Classifier Learning based<br>🟪 Geo-location/GPS based | ⚫ 🟢⚫<br>🟢⚫<br>🟡<br>🔴<br>🔴🟢🟡<br>🟤⚪🟠<br>🔴🟤🟡⚫🟠<br>🟤⚪🟠<br>⚪<br>🟤🟡<br>⚪<br>🩷🟢⚪<br>🟪<br>🔴<br>🔴🟢⚫<br>⚪<br>⚪<br>🔵<br>🟦<br>🔵<br>⚪<br>🟦⚪<br>⚪<br>🔴🟤⚫🟡<br>🟪 | Brown and Lowe (2005)<br>Snavely et al. (2006, 2008a)<br>Snavely et al. (2008b)<br>Chum and Matas (2010)<br>Agarwal et al. (2009)<br>Havlena et al. (2009)<br>Frahm et al. (2010)<br>Gherardi et al. (2010)<br>Sinha et al. (2010)<br>Havlena et al. (2010)<br>Olsson and Enqvist (2011)<br>Crandall et al. (2011)<br>Lou et al. (2012)<br>Cao and Snavely (2012)<br>Wu (2013)<br>Chatterjee and Govindu (2013)<br>Moulon et al. (2013)<br>Jian et al. (2014)<br>Hartmann et al. (2014)<br>Havlena and Schindler (2014)<br>Wilson and Snavely (2014)<br>Shah et al. (2014)<br>Shah et al. (2015)<br>Bhowmick et al. (2014)<br>Schönberger et al. (2015) |
| **Step 2**<br>Find Connected<br>Components | 🟫 From Approximate Match-graph<br>🟩 From Feature-based Match-graph<br>🟨 Find Optimal Connected Set | | |
| **Step 3**<br>Pairwise Feature<br>Matching | 🟦 Optimized Search Structures<br>🟦 Search-space Reduction<br>🟦 Dimensionality Reduction | | |
| **Step 4**<br>SFM Reconstruction<br>of Components | ⚫ Incremental Methods<br>⚫ Global Methods<br>⚪ Hierarchical Methods | | |
| **Step 5**<br>Merge Reconstructed<br>Components | 🟧 Using Common 3D Points<br>🟨 Using Link Images | | |

Table 1: A high-level division of the large-scale reconstruction pipeline into five sub-tasks. The methods and algorithms proposed for each of these subtasks are divided into broad categories. For better understanding of the prior works, key papers are color coded for the categories based on the primary contribution (for methods papers) and based on the utilized techniques (for the pipeline papers). This visualization is best viewed in color.

## 2. Background and Related Work

Recovering structure and motion from multiple images is a long studied problem in computer vision. Early efforts to solve this problem were mostly algebraic in nature, with closed form, linear solutions for two, three, and four views. Hartley and Zisserman (2003) provide a comprehensive account of these now standard techniques. For multi-image sequences with small motions, factorization based solutions were proposed by Tomasi and Kanade (1992) and Sturm and Triggs (1996). Algebraic methods are fast and elegant but sensitive to noisy feature measurements, correspondences, and missing features. Another class of algorithms took a statistical approach and iteratively solved the reconstruction problem by minimizing the distance between the projections of the 3D points in images and feature measurements ("reprojection error") using non-linear least squares technique (Szeliski and Kang, 1993; Taylor et al., 1991). These methods are robust to noise and missing correspondences but computationally more expensive than linear methods. The joint optimization of all camera parameters and 3D points by minimization of the reprojection error is now commonly referred to as Bundle Adjustment (Triggs et al., 2000) which has been a long studied topic in the field of Photogrammetry. Advances in robust feature detection and matching (Lowe, 2004) and sparse bundle adjustment made the structure from motion techniques applicable to unordered photo collections (Brown and Lowe, 2005). Snavely et al. (2006) presented the first system for large-scale 3D reconstruction using the incremental SFM algorithm on Internet photo collections. Since then, many efforts have been made to push the state of the art.

There are two main tasks involved in a typical reconstruction pipeline, (i) match-graph construction - that computes pairwise geometric constraints between the image pairs, and (ii) structure-from-motion reconstruction - that recovers a globally consistent structure from the match-graph. However, in the context of large-scale reconstruction, often these tasks are further divided into sub-taks. The most commonly used pipeline for large-scale SFM can broadly be divided into five sub-tasks as depicted in Table 1.

Match-graph construction begins with a filtering step that identifies the images that potentially have visual overlap (step 1). For city-scale reconstructions, multiple connected-components that can be reconstructed independently are identified from the potential image connections (step 2). For each of the connected-components, a match-graph (or a view-graph) is constructed by performing pairwise feature matching for all directly connected nodes and by verifying the matches based on two-view geometric constraints (step 3). Step 2 and step 3 are sometimes performed in a reverse order, i.e. the connected-components are iden-tified after feature matching. Each connected-component is reconstructed from the pairwise correspondences using structure from motion (step 4) and finally merged into a single reconstruction (step 5). We now explain each of these steps and discuss the related literature in the remainder of this section.

*2.1. Selecting Image Pairs to Match*

Large-scale image collections often contain images that do not capture the structure of interest. Also, a large number of good images do not match with the majority of the other images, as they capture different parts of the structure and have no visual overlap. With tens of thousands of features per image, the cost of pairwise feature matching is non-trivial. Hence, exhaustively matching features between all pairs of images ($O(n^2)$) would result in a wasted effort in performing expensive feature matching between a large number of unmatchable images. Due to this, most large-scale pipelines first attempt to identify the image pairs that can potentially match using computationally cheaper methods.

Many methods use the global similarity between two images as a measure of matchability. Bag of words (BoW) models and vocabulary tree (VOC-tree) based image retrieval techniques are popularly used in SfM context to identify image pairs with potential overlap (Agarwal et al., 2009; Bhowmick et al., 2014). To improve efficiency of this retrieval based scheme, Chum and Matas (2010) employ a min-hashing based technique. To improve quality of retrieved results, Lou et al. (2012) integrate a relevance feedback based approach. Frahm et al. (2010) use global image features (GIST) to cluster images based on their appearance and select an iconic image for each valid cluster. Images within a cluster are only matched with the iconic image and not exhaustively. Iconic images across clusters are evaluated for similarity using VOC-tree retrieval or geo-tags. Crandall et al. (2011) also use geo-tags/GPS and VOC-tree based similarity to reduce the number of image pairs to match.

Another class of methods use learning techniques to evaluate whether an image pair would match. Cao and Snavely (2012) use discriminative learning on BoW histograms and train an SVM classifier to predict matchability. Recently, Schönberger et al. (2015) introduced a pairwise geometry encoding scheme that quantifies distribution of location and orientation changes between an image pair based on feature correspondences and train a random forest classifier for prediction.

A preemptive matching (PM) scheme to quickly eliminate non-promising image pairs was proposed by Wu (2013). Preemptive matching examines matches between a few (100) high-scale features of an image pair and considers the pair for full matching only if 2-4 matches are found among these features.

*2.2. Finding Connected-components*

The image connections found in step 1 define an approximate match-graph with edges between nodes corresponding to matchable pairs. A connected-component is found by performing a depth first search on this approximate match-graph and later pruned by feature matching and geometric verification (Crandall et al., 2011; Bhowmick et al., 2014; Frahm et al., 2010). Some other pipelines perform pairwise feature matching (step 3) first to compute a geometrically verified match-graph (or a view-graph) and then find connected-components (Agarwal et al., 2009; Snavely et al., 2008a).

Some methods also propose to make the connected-components sparser to improve the efficiency of the SFM step. Especially for incremental SFM with $O(n^4)$ cost, the improvement is significant. Snavely et al. (2008b) compute a skeletal-graph (or skeletal-set) by finding a maximum leaf T-spanner in the image graph such that the uncertainty of pairwise reconstructions is minimized without loss of image connectivity. Skeletal-set computation is done after pairwise feature matching and geometry estimation. To avoid this, Havlena et al. (2010) pose the problem of finding a sparse but connected graph as that of finding a minimum connected dominating set (min CDS) from a pairwise image similarity matrix computed using the BoW model.

*2.3. Pairwise Feature Matching*

Features of two images are matched by computing $L_2$-norm between the corresponding descriptors and finding the closest feature as a candidate match. Since, $L_2$-norm in descriptor space is not meaningful by itself to indicate a match, the candidate match is verified by performing a ratio-test, i.e. by verifying that the ratio of distances of the query feature from its top-two neighbors in the target image is below a threshold.

High-resolution images of structure have tens of thousands of point features. Without using massively multi-threaded hardware like GPUs, exhaustively comparing features between two images is computationally prohibitive ($O(m^2)$ for m features per image) even after reducing the image pairs. Hence, it is common to use approximate nearest-neighbor search ($O(m \log m)$) using Kd-trees for feature matching. To further improve the efficiency of feature search, Jian et al. (2014) proposed a cascade hashing based approach.

Alternatively, the efficiency of feature matching can be improved by reducing the search space. Hartmann et al. (2014) observe that many features that occur repetitively participate in matching but are later discarded by ratio-test. To eliminate such features from the matching pool, they train random forest classifiers to predict a feature's matchability. Havlena and Schindler (2014) suggest that if features are quantized into a very large vocabulary, the quantization would be sufficiently fine to assume that features from multiple images belonging to the same visual word are matches. However, they discard words that contain many features from the same image as they are either noisy features or features that occur on repetitive structures. Hartmann et al. (2014) and Havlena and Schindler (2014) discard matchable features on repetitive structures as they are traditionally rejected during ratio-test. To avoid this, in Shah et al. (2015), we presented a two-stage geometry aware scheme that leverages coarse epipolar geometry to reduce the number of candidate features to match and also produces denser correspondences by retaining good correspondences on repetitive structures.

*2.4. Reconstruction of a Connected Component*

Given the match-graph/view-graph for a connected component, reconstruction can be performed by various structure from motion pipelines. Most SFM reconstruction techniques can be categorized into, (i) Incremental SFM, (ii) Global SFM, and (iii) Hierarchical SFM.

Incremental SfM (Brown and Lowe, 2005; Snavely et al., 2006) reconstructs the cameras and points starting with a good seed image pair. The reconstruction grows incrementally by adding a few well connected images, estimating the camera parameters, and triangulating feature matches. To avoid drift accumulation, this is followed by a global bundle adjustment (BA) which refines camera poses and 3D point positions. The complexity of the incremental SFM is $O(n^4)$ due to repeated BA. To improve the efficiency of this step, many methods propose fast approximations of the sparse bundle adjustment and/or exploit many-core architectures to parallelize it (Agarwal et al., 2009, 2010b; Wu et al., 2011; Byröd and Åström, 2010). Wu (2013) leverages highly parallel GPU architecture for an optimized pipeline - VisualSFM for incremental SFM. In this pipeline, progressively fewer iterations of BA are performed as more images are added, owing to the observation that the recovered structures become stable as they grow larger.

Another class of methods can be classified as global SFM methods as they aim to reconstruct all images at once as opposed to a sequential solution. Sinha et al. (2010) estimate global camera rotations from pairwise rotations and vanishing points. Olsson and Enqvist (2011) use rotation averaging in RANSAC framework for global rotation estimation. Crandall et al. (2011) propose a discrete-continuous optimization method, Chatterjee and Govindu (2013) use lie-algebraic averaging for global rotation estimation. Once the rotations are known, SFM boils down to solving a linear problem of estimating camera translations and 3D structure. Wilson and Snavely (2014) go a step further and also estimate global translations by solving for 1D ordering in a graph problem. Due to averaging of pairwise motions, global methods perform poorly when the pairwise geometries are inaccurate, or there are fewer pairwise geometries to average. Recently, Sweeney et al. (2015) proposed a view-graph optimization scheme that works for uncalibrated image sets and also improves the robustness of global methods to handle inaccuracies in pairwise geometries.

Havlena et al. (2009) and Gherardi et al. (2010) proposed hierarchical methods for SFM that attempt to avoid fully sequential reconstruction typical to incremental SFM methods without using global estimations. Havlena et al. (2009) finds candidate image triplets using visual words for atomic three image reconstructions and merges them into a larger reconstruction. Gherardi et al. (2010) organizes the images into a balanced tree using agglomerative clustering on the match-graph and builds a larger reconstruction by hierarchically merging the separately reconstructed clusters. In Shah et al. (2014), we proposed a multistage approach for SFM that first reconstructs a coarse global model using a match-graph of a few high-scale features and enriches it later by simultaneously localizing additional images and triangulating additional points. Leveraging the known geometry of the coarse model allows the later stages of the pipeline to be highly parallel and efficient for component-level reconstruction.

*2.5. Merging reconstructed components*

The connected components of the match-graph are independently reconstructed using methods discussed before and later merged into a single reconstruction. Many pipelines merge multiple sub-models by finding the common 3D points across the models and by robustly estimating a similarity transformation using ARRSAC/ RANSAC/ MSAC (Frahm et al., 2010; Raguram et al., 2011; Havlena et al., 2009; Gherardi et al., 2010). These merging methods mainly differ in their ways of identifying common 3D points. Bhowmick et al. (2014), while dividing the match-graph into multiple components ensure that common images exist between two components and estimate the similarity transform between two models by leveraging the pairwise epipolar geometry of the link images. Recently, Cohen et al. (2015) presented a combinatorial approach for merging visually disconnected models of urban scenes.

*Our work in the large-scale context:* The methods discussed in this paper contribute to the steps 3 and 4 of the large-scale pipeline and hence we evaluate them against works within the related contexts in section 7. Our multistage framework is mainly an alternate staging of existing SFM techniques. Independent improvements in each of these techniques can be incorporated into our framework for its continual improvement.

## 3. Overview of Multistage SFM Algorithm

The flow of our algorithm is depicted in Figure 1. We begin with a set of roughly-connected images that represent a single a monument or geographic site. Appearance techniques and geotags can be used to obtain such image components from larger datasets as explained in section 2. Alternatively, images of a site may be captured or collected specifically for image based modeling, e.g. for digital heritage applications. We first extract SIFT features from these images and sort them based on their scales. Our algorithm then operates in the following main stages.

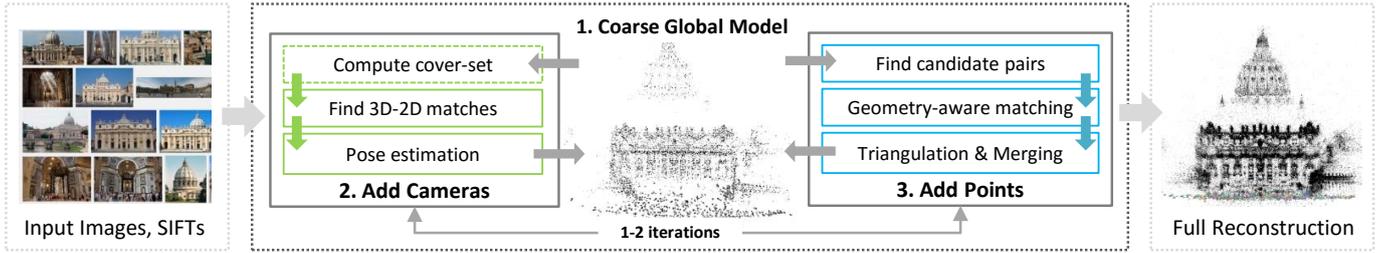

Figure 1: Flow of our multistage algorithm. Given images of a component, in stage 1, we match a small fraction of the image SIFTs and recover a coarse but global model of the scene using any robust SFM method. In stage 2, camera poses for the the images un-registered in the first stage are estimated using fast 3D-2D correspondence search based localization. In stage 3, the unmatched features of the localized images are matched with a set of candidate images using geometry-aware matching and triangulated to produce the final model. Stages 2 and 3 are highly parallel and do not require bundle adjustment. These stages can be repeated for more iterations if needed.

*Coarse model estimation* In this stage, we match a few high-scale SIFTs of given images and estimate a coarse global model of the scene using any robust SFM method. Given the coarse model, the remaining reconstruction problem is formulated as stages of simultaneously adding remaining cameras and then simultaneously adding remaining points to this model. This breaks the essential sequentiality of incremental SFM and provides a mechanism to get faster results by using more compute power.

*Adding cameras* This stage estimates camera poses for images that could not be registered to the coarse model recovered from high-scale SIFTs. We use 3D-2D matching based image localization for this. (Sattler et al., 2011; Li et al., 2010; Irschara et al., 2009; Choudhary and Narayanan, 2012). Since camera pose is estimated using direct 3D-2D correspondences between given image and the model, images can be localized independently of each other.

*Adding points* This stage enriches coarse reconstruction by matching and triangulating remaining SIFT features. We exploit the camera poses recovered in earlier stages in two ways. First, it reduces the pairwise image matching effort by matching only the image pairs with covisible 3D points in the coarse model. Second, it leverages the epipolar constraints for fast geometry-aware feature search. Our point addition stage recovers denser point clouds as guided matching helps to retain many valid correspondences on repetitive structures. Such features are discarded in unguided matching since ratio-test is performed before geometric verification.

The models reconstructed using our multistage approach converge to full-models reconstructed using all features and traditional pipelines in 1-2 iterations of the camera and point addition stages. Since we begin with a global coarse model, our method does not suffer from accumulated drifts (for datasets observed so far), making incremental bundle adjustment optional in later stages of our pipeline. Please see section 7 for a detailed discussion of these results.

*3.1. Terminology*

We borrow and extend the terminology used in previous papers (Li et al., 2010; Choudhary and Narayanan, 2012; Snavely et al., 2008a). Let $\mathbb{I} = \{I_1, I_2, ..., I_n\}$ be the set of input images. Each image $I_i$ contains a set of *features* $F_i = \{f_k\}$, each feature represents a 2D point and has a 128-dim SIFT descriptor associated with it. Let $\mathbb{M} = \langle \mathbb{P}, \mathbb{C} \rangle$ denote the 3D model which we wish to approximate, where $\mathbb{P} = \{P_1, P_2, \cdots, P_m\}$ is the set of 3D points and $\mathbb{C} = \{C_1, C_2, \cdots, C_n\}$ is the set of cameras. The coarse model is denoted as $\mathbb{M}_0$. Subsequently in $i^{th}$ iteration, the models after the camera addition (localization) and point addition stages are denoted as $\mathbb{M}_i^l$ and $\mathbb{M}_i$ respectively.

An image $I_i$ gets upgraded to a camera $C_i$ when its projection matrix is estimated, giving a one-to-one mapping between images and cameras. We use the terms camera $C_i$ and image $I_i$ interchangeably as per the context. A feature $f$ gets upgraded to a point $P$ when its 3D coordinates are known. However, corresponding features are projections of the same point in different cameras giving a one-to-many mapping that we define as the $Track$ of a point. $Track(P_k)$ would map point $P_k$ to a set $\{(C_i, f_j)\}$, where the features $f_j$s are projections of the point $P_k$ in cameras $C_i$. Similar to Snavely et al. (2008a), we define two mappings $Points(.)$ and $Cameras(.)$. $Points(C_i)$ indicates a subset of $\mathbb{P}$ consisting of all points visible in camera $C_i$ and $Cameras(P_j)$ indicates a subset of $\mathbb{C}$ consisting of all cameras that see point $P_j$.

We explain each step of our multistage approach with algorithmic and implementation details in the following sections.

## 4. Coarse Model Estimation

The coarse global model is estimated by SFM reconstruction of high-scale feature match-graph. Any robust SFM method can be used for this reconstruction. We experimented with both incremental and global methods for coarse reconstruction. These experiments are discussed in detail in section 7. In this section, we focus on the feature selection criteria.

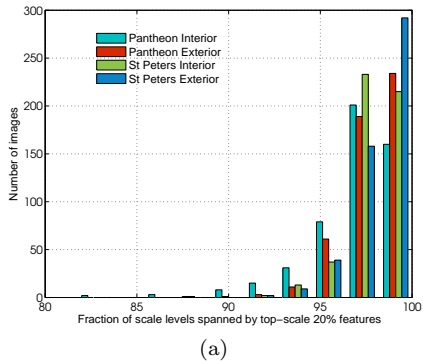
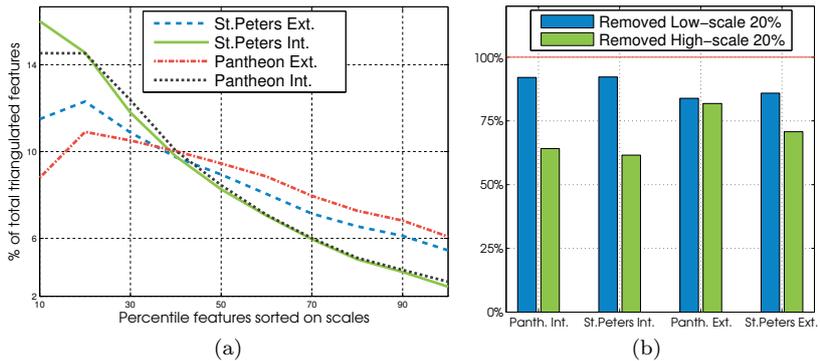

Figure 2: Histograms representing the fractions of scale levels spanned by the top-scale 20% features for sets of 500 images each randomly sampled from four datasets. High-scale features cover many scale-levels.

Figure 3: Analysis of triangulated features by scales in reconstructed models: (a) illustrates the distribution of triangulated features vs. their percentile scale rank, (b) illustrates the effect of removing high vs. low scale features on total number of triangulated points. These plots indicate that high-scale features participate more commonly in triangulated feature tracks and are clearly more important than low-scale features for reconstruction.

For match-graph construction in this stage, we select only $\eta\%$ features from all images in descending order of their scales. One should note that this is very different from extracting features from down-sampled images or picking random features. There are two main reasons why we favor higher-scale features for reconstruction: (i) Due to successive gaussian blurring applied to create a scale-space signal, fewer and fewer features are detected at higher scales. Hence, the selected coarse features span across many scales of the scale-space structure. Figure 2 shows the histograms of fractions of scale-levels spanned by the top-scale 20% features of randomly sampled 500 images for four datasets. It can be seen that for most of the images across all datasets, more than 90% of the scale-levels are spanned by the selected coarse features, indicating broad coverage; (ii) Features detected from the top scale signals correspond to more stable structures in the scene as compared to the features detected at high-resolution bottom scales which are more susceptible to change with minor variations in the imaging conditions. Due to these two reasons, we consider high-scale features both reliable and sufficient for coarse image matching and geometry estimation.

The latter observation is empirically verified by analyzing the distribution of features by their scales in different models reconstructed using a standard structure from motion pipeline - Bundler. Figure 3a shows the distribution of reconstructed features vs. their percentile rank by scale for four models. Higher-scale points clearly are part of more 3D points. The area under the curve is high for $\eta$ value of 10–30. Choosing these features for coarse model reconstruction would enable us to recover many 3D points. Figure 3b shows the number of 3D point tracks that would survive if the top 20% and bottom 20% features by scale are removed from the tracks. The high-scale features are clearly more important than the low-scale ones, as more points are dropped when they are removed. It also indicates that high-scale features not only match well but they also match more frequently to other features of higher scales. We also performed experiments with matchability prediction (Hartmann et al., 2014) for feature selection but found the scale-based selection strategy to be more effective for coarse reconstruction. To show this, we construct match-graphs using 20% features, (a) selected based on scales, (b) selected using matchability prediction. Table 2 shows number of matches and number of connected pairs for four datasets using both selection strategies. The scale-based selection produces similarly or more connected match-graphs as compared to matchability prediction based selection. Details of these datasets are given in Table 3 in section 7.

| dataset | Scale-based Selection | | Matchability Prediction | |
|---|---|---|---|---|
| | #matches | #pairs | #matches | #pairs |
| PTI | 1.44M | 30K | 1.10M | 29K |
| PTE | 13.09M | 152K | 5.3M | 127K |
| SPI | 3.81M | 64K | 1.83M | 50K |
| SPE | 11.4M | 223K | 8.3M | 239K |

Table 2: Number of matches and connected image pairs (pairs with > 16 matches) for two feature selection strategies

We performed various experiments to see the effect of $\eta$ on completeness of reconstruction and runtime. We conclude that selecting 20% high-scale for initial matching offers an optimum trade-off between connectivity and matching efficiency for images with $10K - 30K$ features. The complexity for Kd-tree based pairwise feature matching is $O(n^2 m \log m)$, for $n$ images and $m$ features per image. Most literature on SFM ignores $m$, assuming it to be a small constant. However, typical images have tens of thousands of features and $m$ does have an impact on runtime in practice. Since we use only $\eta\%$ of features, feature matching is nearly $100/\eta$ times faster for components of $\sim 1000$ images. Fewer features also have a considerable advantage on practical runtime of bundle adjustment during reconstruction (see details in section 7).

To further improve the efficiency, we adopt a hybrid matching scheme inspired by preemptive matching. Here, the first 10% of high-scale query features are matched with 20% features of the reference image. Then, the next 10% query features are matched only if sufficient matches are found ($> 4$) in the first batch. Matching can be terminated early if sufficient matches for geometry estimation are found. Please note that for images with very few features ($< 1000$), we simply use all features for matching. We use the Approximate Nearest Neighbour (ANN) library (Arya et al., 1998) for Kd-tree based feature matching on CPU (Arya et al., 1998) and SIFTGPU library (Wu, 2007) for global feature matching on GPU.

We denote the recovered model as $\mathbb{M}_0 = \langle \mathbb{C}_0, \mathbb{P}_0 \rangle$, where $\mathbb{C}_0$ is the set of pose estimated (localized) images and $\mathbb{P}_0$ is the set of 3D points mapped to triangulated feature tracks. The model $\mathbb{M}_0$ is *coarse* but *global*. That is, as compared to the full models reconstructed using all features, the 3D points in $\mathbb{P}_0$ are sparser but $\mathbb{C}_0$ covers most of the cameras. In our experiments, $\mathbb{M}_0$ contained 80%-100% of the cameras of the full construction and roughly $\eta$% of the 3D points. The coarse model, however, contains enough information to successfully add remaining cameras and points in subsequent stages.

## 5. Adding Cameras to the Model

After the reconstruction of the coarse model $\mathbb{M}_0$ in the first stage, some of the images would remain unregistered and a large number of features in all images would remain yet to be matched and triangulated. In this stage, we enrich the model $\mathbb{M}_0$ by registering the remaining images to the model. This stage is later repeated after the point addition stage if needed.

Registering a new image to an existing SFM model is known as the localization problem. For localization, it is necessary to establish correspondences between 3D points in the model and 2D features of the image to be localized. Once reliable and sufficient 3D-2D matches are established, the camera pose can be estimated using PnP solvers. Since, a global SFM model is known, localization of each image can be done independently of others. Unlike the traditional incremental SFM process where images can only be added to a growing reconstruction in a sequential manner, the coarse model allows us to localize all unregistered images simultaneously, and in parallel. Many methods have been proposed for efficient image localization (Li et al., 2010; Choudhary and Narayanan, 2012; Sattler et al., 2011, 2012; Irschara et al., 2009). These methods mainly differ in their strategies for 3D-2D correspondence search. We have experimented with three different strategies for correspondence search in our pipeline.

*Direct 3D-2D matching* In Shah et al. (2014), we used a naïve, direct 3D-2D matching approach. In this method, each 3D point $P_i$ in the model is represented by the mean SIFT descriptor of all features in its track, $Track(P_i)$ and queried into the Kd-tree of all feature descriptors of the image to be localized. A match is declared by the ratio-test. With this method for correspondence search, localization takes around 1 to 5 seconds to localize a single image, for models of about 100K-200K 3D points.

*Active correspondence search* In this paper, we replace our direct 3D-2D matching approach with the state-of-the-art active correspondence search proposed by Sattler et al. (2012). In this technique, the pitfalls of both 3D-to-2D search and 2D-to-3D search are avoided by a combined approach. Moreover, the technique is made efficient by incorporating prioritized search based on visual words. This localization technique is significantly faster and superior in quality as compared to our previous technique. Localizing an image using active correspondence search takes between 0.5 to 1 second. The pre-processing steps for this search involve computing mean SIFT descriptors and visual word quantization which are easy to parallelize.

*Ranked 2D-2D matching* Many image pairs with coarse feature matches do not participate in the SFM step for coarse model estimation due to either insufficient matches or inliers. Nevertheless, matches between the coarse features of an image pair offer an important cue that the images could visually overlap. We leverage this cue and propose a ranked 2D-2D matching scheme for 3D-2D correspondence search when active search fails. Given an unlocalized image, we find the top-K localized images sorted on the number of coarse feature matches they share with the unlocalized image. For each of the K localized images, we find the subset of 2D features that participate in tracks of 3D points in the current model and use these feature descriptors as proxies for 3D-2D matching. We create a Kd-tree of all features in the unlocalized image, query the subset of 2D features of the nearby localized images into this Kd-tree, and verify matches by ratio-test, thereby establishing correspondences ($> 16$) between the parent 3D points and the 2D features in the unlocalized image.

While the coarse model is typically small and localization is fast in the first iteration, the model after the first point addition stage gets heavy in 3D points for efficient localization in the later iterations. To avoid this, we use set cover representation of a 3D model, if it contains $> 100K$ 3D points. The set cover of a model is a reduced set of points that cover each camera at least $k$ (300-500) times (Li et al., 2010). Upon obtaining sufficient number of 3D-2D matches, RANSAC based pose estimation and non-linear pose refinement are performed, and finally the model is updated with all localized images.

By addition of newly localized cameras, the model $\mathbb{M}_i = \langle \mathbb{C}_i, \mathbb{P}_i \rangle$ upgrades to an intermediate model $\mathbb{M}_i^l = \langle \mathbb{C}_{i+1}, \mathbb{P}_i \rangle$. For each localized camera $C_q$, we have the inlier 3D-2D correspondences $(P_j \leftrightarrow f_k)$. We update all $Track(P_j)$'s to contain $(C_q, f_k)$ after adding each camera $C_q$. The new cameras each have a few points at this stage. More points are added for all pose-estimated cameras in the subsequent point addition stage.

## 6. Adding Points to the Model

The point addition stage updates the model $\mathbb{M}_i^l = \langle \mathbb{C}_{i+1}, \mathbb{P}_i \rangle$ to $\mathbb{M}_{i+1} = \langle \mathbb{C}_{i+1}, \mathbb{P}_{i+1} \rangle$ by triangulating several unmatched features of images in $\mathbb{C}_{i+1}$. The model after first camera addition stage is nearly complete in cameras but consists of points corresponding to only $\eta\%$ high-scale features of localized cameras. After the first point addition step, the model is dense in points. This step is repeated after every round of camera addition to triangulate and merge features of the newly added cameras. This is done to ensure that un-localized cameras can form 3D-2D connections with newly localized cameras too in the upcoming camera addition stage. To accelerate this stage, we leverage the known geometry of the existing model in the following two ways, (i) we use the visibility relations between localized cameras and triangulated coarse features to restrict feature matching to only pairs with sufficiently many co-visible points. (ii) we use the epipolar geometry between the localized cameras to accelerate feature correspondence search. In the following sections, we explain these individual steps in detail.

### 6.1. Finding Candidate Images to Match

Given a set of images of a monument or a site, each image would find sufficient feature matches with only a small fraction of total images; those looking at common scene elements. Ideally we would like to limit our search to only these *candidate* images. We use the point-camera visibility relations of the model $\mathbb{M}_1^l = \langle \mathbb{C}_1, \mathbb{P}_0 \rangle$ to determine whether or not two images are looking at common scene elements.

Let $I_q$ denote the query image and $F_q = \{f_1, f_2, ..., f_m\}$ denote the features that we wish to match and triangulate. Traditionally we would attempt to match the features in image $I_q$ with the features in set of all localized images $I_L$ where, $I_L = \{I_i \mid C_i \in \mathbb{C}_1, C_i \neq C_q\}$. However, we wish to match the features in query image $I_q$ with features in only a few *candidate* images that have co-visible points with image $I_q$. We define the set of all co-visible points between two images $I_i$ and $I_j$ as, $P_{cv}(I_i, I_j) = Points(C_i) \cap Points(C_q)$. Using this visibility relations, we define the set of candidate images for image $I_q$ as, $S_q = \{I_i \mid |P_{cv}(I_q, I_i)| > T\}$ ($T = 8$ for our experiments). We select only top-$k$ candidate images ranked based on the number of co-visible points. Our experiments show it is possible to converge to a full match-graph of exhaustive pair-wise matching even when the number of candidate images $k$ is limited to only 10% of the total images. We find unique image pairs from candidate image sets for all query images and match these pairs in parallel using fast geometry-aware feature matching.

### 6.2. Fast Geometry-aware Feature Matching

Given a query image $I_q$ and its candidate set $S_q$, we use the guided matching strategy to match the feature sets $(F_q, F_c \mid I_c \in S_q)$. In traditional feature matching each query feature in $F_q$ is compared against features in a candidate image using a Kd-tree of features in $F_c$.

Since query image $I_q$ and candidate image $I_c$ both are localized, their camera poses are known. Given the intrinsic matrices $\mathsf{K_q}$, $\mathsf{K_c}$, rotation matrices $\mathsf{R_q}$, $\mathsf{R_c}$, and translation vectors $\mathsf{t_q}$, $\mathsf{t_c}$, the fundamental matrix $\mathsf{F_{qc}}$ between image pair $I_q$ and $I_c$ can be computed as,

$$\mathsf{F_{qc}} = \mathsf{K_q^{-T} R_q [R_c^T t_c - R_q^T t_q]_\times R_c^T K_c^{-1}}. \quad (1)$$

For a query feature point $\mathsf{p_q} = (\mathsf{x_q}\ \mathsf{y_q}\ 1)$ in feature set $F_q$ of image $I_q$ the corresponding epipolar line $\mathsf{l_q} = (\mathsf{a_q}, \mathsf{b_q}, \mathsf{c_q})$ in image $I_c$ is given by $\mathsf{l_q} = \mathsf{F_{qc}} \cdot \mathsf{p_q}$. If $\mathsf{p_q'} = (\mathsf{x_q'}\ \mathsf{y_q'}\ 1)$ denotes the corresponding feature point in image $I_c$ then as per the epipolar constraint $\mathsf{p_q'} \cdot \mathsf{F_{qc} p_q} = 0$, point $\mathsf{p_q'}$ must lie on the epipolar line i.e. $\mathsf{p_q'} \cdot \mathsf{l_q} = 0$. Due to inaccuracies in estimation, it is practical to relax the constraint to $\mathsf{p_q'} \cdot \mathsf{l_q} < \epsilon$. To find the corresponding point $\mathsf{p_q'}$, instead of considering all features in set $F_c$, we limit our search to only those features which are close to the epipolar line $\mathsf{l_q}$. We define the set of candidate feature matches $\mathbf{C}$ as,

$$\mathbf{C} = \{\mathsf{p'} \mid \mathrm{dist}(\mathsf{p'}, \mathsf{l_q}) \leq \mathsf{d}\} \quad (2)$$

$$\mathrm{dist}(\mathsf{p'}, \mathsf{l_q}) = \frac{\mathsf{a_q x' + b_q y' + c_q}}{\sqrt{\mathsf{a_q}^2 + \mathsf{b_q}^2}} \quad (3)$$

We propose a fast algorithm for finding the set of candidate feature and propose an optimal strategy for correspondence search based on the dual nature of epipolar lines.

#### 6.2.1. Finding the Set of Candidate Matches

*Linear search:* In Figure 4 the candidate feature matches (features in set $\mathbf{C}$) are marked by red dots. Finding these candidate matches using linear search would require computing the distances of all features in $F_c$ from line $\mathsf{l_q}$ using equation (3). This search has a time complexity of $O(|F_c|)$.

*Radial search:* Linear search can be approximated by a faster radial search algorithm of logarithmic complexity. In this search, first a Kd-tree of $(\mathsf{x}, \mathsf{y})$ coordinates of features in $F_c$ is constructed. Then $\mathsf{K}$ equidistant points (at distance d) on the epipolar line $\mathsf{l_q}$ are sampled and each of these points is queried into the Kd-tree to retrieve feature points within radial distance d from the sampled point (Muja and Lowe, 2014). In Figure 4b dark green squares on the epipolar line mark the equidistant query points and red circles indicate coverage of *true* candidate matches when radial search is used. If line $\mathsf{l_q}$ intersects image $I_c$ in points $\mathsf{p_A} = (\mathsf{x_A}, \mathsf{y_A})$ and $\mathsf{p_B} = (\mathsf{x_B}, \mathsf{y_B})$ then the

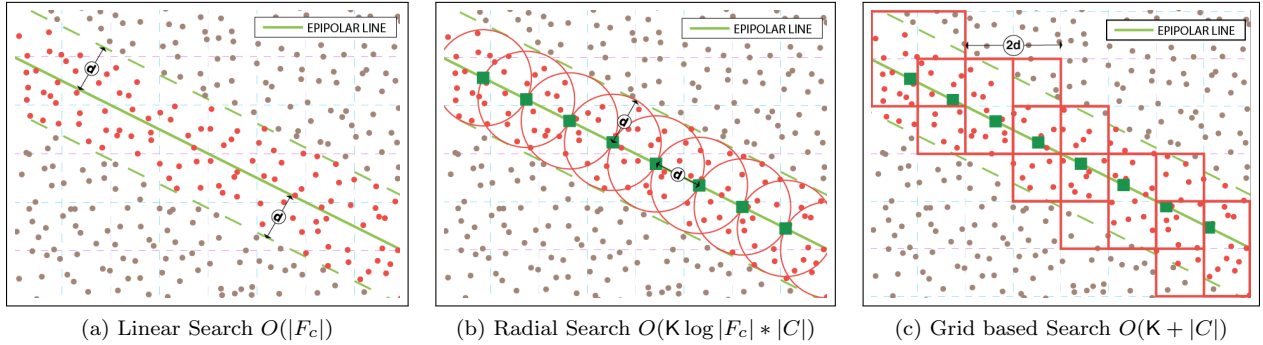

(a) Linear Search $O(|F_c|)$    (b) Radial Search $O(\mathsf{K} \log |F_c| * |C|)$    (c) Grid based Search $O(\mathsf{K} + |C|)$

Figure 4: Illustration of geometry-aware search strategy. Search for points within distance d from epipolar line (shown by red dots) can be approximated by radial search and more efficient grid based search. Red squares in (c) shows the center-most cell of the overlapping grids selected for each equidistant points along the epipolar line.

coordinates $(\mathsf{x_k}, \mathsf{y_k})$ of the equidistant points are given by,

$$\mathsf{x_k} = \frac{\mathsf{k} \cdot \mathsf{x_A} + (\mathsf{K} - \mathsf{k}) \cdot \mathsf{x_B}}{\mathsf{K}}, \quad \mathsf{k} = 0, 1, 2, \cdots, \mathsf{K} \quad (4)$$

$$\mathsf{y_k} = \frac{\mathsf{k} \cdot \mathsf{y_A} + (\mathsf{K} - \mathsf{k}) \cdot \mathsf{y_B}}{\mathsf{K}}, \quad \mathsf{k} = 0, 1, 2, \cdots, \mathsf{K} \quad (5)$$

where $\mathsf{K} = \sqrt{(\mathsf{x_B} - \mathsf{x_A})^2 + (\mathsf{y_B} - \mathsf{y_A})^2}/\mathrm{d}$.

The number of leaf nodes visited in the Kd-tree depends on the number of features to retrieve ($|C|$). The complexity of radial search is $O(\mathsf{K} \cdot \log |F_c| * |C|)$, $\mathsf{K} \ll |F_c|$.

*Grid-based search:* We further optimize the candidate search by using a grid based approach. We first divide the target image $I_c$ into four overlapping grids of cell size 2d × 2d and overlap of d along x, y and x-y directions, as shown by dotted lines in Figure 4c. We then bin all feature points of the set $F_c$ into cells of the overlapping grids based on their image coordinates. Each feature point (x, y) would fall into four cells, coordinates of centers of these cells are given by,

$$\mathsf{x_{c1}} = \lfloor \frac{\mathsf{x}}{2\mathrm{d}} \rfloor \cdot \mathrm{d} + 2\mathrm{d}, \quad \mathsf{y_{c1}} = \lfloor \frac{\mathsf{y}}{2\mathrm{d}} \rfloor \cdot \mathrm{d} + 2\mathrm{d} \quad (6)$$

$$\mathsf{x_{c2}} = \lfloor \frac{\mathsf{x}}{2\mathrm{d}} - \frac{1}{2} \rfloor \cdot \mathrm{d} + 2\mathrm{d}, \quad \mathsf{y_{c2}} = \lfloor \frac{\mathsf{y}}{2\mathrm{d}} \rfloor \cdot \mathrm{d} + 2\mathrm{d} \quad (7)$$

$$\mathsf{x_{c3}} = \lfloor \frac{\mathsf{x}}{2\mathrm{d}} \rfloor \cdot \mathrm{d} + 2\mathrm{d}, \quad \mathsf{y_{c3}} = \lfloor \frac{\mathsf{y}}{2\mathrm{d}} - \frac{1}{2} \rfloor \cdot \mathrm{d} + 2\mathrm{d} \quad (8)$$

$$\mathsf{x_{c4}} = \lfloor \frac{\mathsf{x}}{2\mathrm{d}} - \frac{1}{2} \rfloor \cdot \mathrm{d} + 2\mathrm{d}, \quad \mathsf{y_{c4}} = \lfloor \frac{\mathsf{y}}{2\mathrm{d}} - \frac{1}{2} \rfloor \cdot \mathrm{d} + 2\mathrm{d} \quad (9)$$

Given a query point $\mathsf{p_q}$, we find its epipolar line $\mathsf{l_q}$ and the equidistant points $(\mathsf{x_k}, \mathsf{y_k})$ as per equations (4) and (5). For each of the equidistant points on the epipolar line, we find the four overlapping cells that contain this point and find its Cartesian distance from centers of the four cells. We select the center most cell for each point and accumulate all feature points binned into these cells to obtain an approximate set of candidate matches $\mathbf{C'}$. Red squares in Figure 4c indicate the coverage of true candidate matches in set $\mathbf{C}$ by grid based approximate search. In practice, we use a slightly larger grid size to account for misses due to grid approximation. Since feature points are binned only once per image, searching candidate matches involves finding center-most cells for $\mathsf{f}K$ points along the line and accumulating $|C'|$ candidate points, leading to runtime complexity of $O(\mathsf{K} + |C'|)$.

*6.2.2. Further optimization*

To finalize a match from candidate set $\mathbf{C'}$, a Kd-tree of feature descriptors in $\mathbf{C'}$ is constructed, closest two features from the query are retrieved, and ratio-test is performed. The number of candidate feature matches $|\mathbf{C'}|$ is a small fraction of total points $|F_c|$ (typically 200:1 in our experiments). Since the top two neighbors are searched in a much smaller Kd-tree of size $|\mathbf{C'}|$, geometry-aware search reduces the operations required for two-image matching from $(|F_q| \log |F_c|)$ to $(|F_q| \log |\mathbf{C'}|)$, with an overhead of constructing a small Kd-tree of size $|\mathbf{C'}|$ for each query feature.

To reduce the overhead of redundant Kd-tree construction, we exploit the dual nature of epipolar lines; i.e. for all points that lie on line l in image $I_c$, their corresponding points must lie on the dual line l' in image $I_q$. We use this property, to group the query points in $I_q$ whose epipolar lines intersect the boundaries of $I_c$ in nearby points (within 2 pixels) and search for matches group by group. Since all feature points in a group have the same epipolar line and hence the same set of candidate matches, we avoid redundant construction of the small Kd-tree of size $|\mathbf{C'}|$ for points in a group.

*6.2.3. Other Advantages*

Apart from being faster than geometry-blind global correspondence search, there are additional advantages of our grid-based geometry-aware search strategy.

*Denser correspondences :* Ratio-test compares the distance of a query feature from its closest neighbor (candidate) to its second closest neighbor in the target image. If the ratio of distances is below a threshold then the candidate is declared a match. The assumption is that features in an image are randomly distributed in descriptor space. In absence of a true match, the candidate would be an arbitrary feature and the best distance would not be significantly better than the second-best distance; leading to a ratio close to 1 (Lowe, 2004). However, the assumption of randomly distributed features does not hold true for images with repetitive structures such as windows, arches, pillars, etc. in architectural images. Ratio-test punitively rejects many correspondences for features on such repeating elements due to similar appearance. In geometry-aware matching, each query feature is compared only against a small number of features that lie close to the corresponding epipolar line. This simple trick reduces the false rejections on repetitive structures significantly by avoiding the duplicates from comparison and recovers denser matches as compared to global matching (see Figure 5).

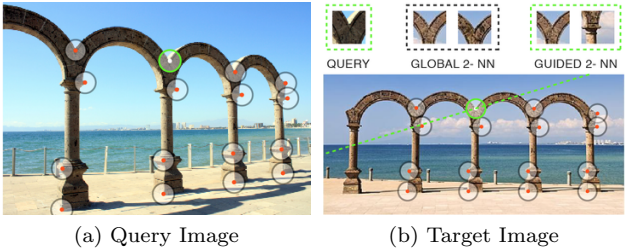

| (a) Query Image | (b) Target Image |

Figure 5: Global vs. geometry-aware matching for repetitive structures: correct match is detected for both cases, but 2-NN for global matching would fail the ratio test due to similarity.

*Easy Parallelization:* While it is straight-forward to parallelize feature matching across image pairs, most approaches that parallelize feature matching within an image pair use exhaustive descriptor comparisons. For high-dimensional data points, Kd-tree based approximate methods are difficult to parallelize due to hierarchical nature of the search. However, the epipolar line based grouping and grid-based candidate search can both be easily distributed on parallel platforms like GPU. We have explained the major steps of our GPU algorithm in appendix A. We show significant speedup (upto 9x) of our GPU matching as compared to state-of-the-art GPU feature matching in section 7.

### 6.3. Triangulation and Merging

After pairwise image matching is performed, we form tracks for features in a query image by augmenting matches found in all candidate images and triangulate these feature tracks using a standard least mean squared error method. We perform this operation independently for all images. This would typically results in duplication of many 3D points because a triangulated feature pair $(C_i, f_k) \leftrightarrow (C_j, f_l)$ for image $C_i$ would also match and triangulate in reverse order for image $C_j$. Also, since we limited our matching to only candidate images, the longest track would only be as long as the size of the candidate set. We solve both of these problems in a track merging step. Our track merging step is similar to Snavely et al. (2006) and uses the standard sequential depth-first-search (DFS) algorithm to find connected-components. It is possible to substitute our sequential implementation with a faster multi-core CPU or GPU implementation.

## 7. Results and Discussion

We evaluate our method on several datasets and discuss quantitative results and run-time performances. First, we show a stand-alone evaluation of the geometry-aware matching as it is the core components of the multistage approach. We show qualitative, quantitative, as well as run-time advantages. Second, we discuss in detail the reconstruction statistics of each stage of the multistage pipeline and compare the results with baseline models. Third, we discuss and compare the run-time performance of various componenets of our pipeline and other standard methods. Finally, we conclude this section with a discussion on limitations and future works.

Table 3 provides details of the different datasets used for evaluating our method. Please note that not all datasets are used across all experiments. We refer to the datasets by the labels given in the second column throughout this section for brevity.

| dataset | label | #images | #feat (avg.) |
| --- | --- | --- | --- |
| Notre Dame Paris (subset)[1] | NDP | 99 | 21K |
| Tsinghua School Building[2] | TSB | 193 | 26K |
| Barcelona National Museum[3] | BNM | 191 | 18K |
| Pantheon Interior[4] | PTI | 587 | 9K |
| Pantheon Exterior[4] | PTE | 782 | 13K |
| St. Peters Interior[4] | SPI | 953 | 15K |
| St. Peters Exterior[4] | SPE | 1155 | 17K |
| Hampi Vitthala Temple[5] | HVT | 3017 | 39K |
| Cornell Arts Quad[6] | AQD | 6014 | 18K |

Table 3: Datasets used in various experiments.

### 7.1. Evaluation of Geometry-aware Matching

To evaluate the effectiveness of geometry-aware matching, we perform 3D reconstruction using Bundler from match-graphs constructed using Kd- tree based matching, cascade hashing based matching (CascadeHash), and two-stage geometry-aware matching methods for three small datasets (NDP, TSB, and BNM in Table 3). Since geometry-aware method depends upon the coarse epipolar geometry between the image pair, we first match the high-scale 20% features using Kd-tree based matching and estimate pairwise fundamental matrices from the initial matches using DLT and RANSAC. The estimated fundamental matrices are then directly used for geometry-aware matching of the unmatched features as explained in subsection 6.2. For this set of experiments, SFM on coarse match-graph is not performed.

---

[1] Snavely et al. (2006) http://phototour.cs.washington.edu/datasets/
[2] Jian et al. (2014) http://vision.ia.ac.cn/data/index.html
[3] Cohen et al. (2012) https://www.inf.ethz.ch/personal/acohen/papers/symmetryBA.php
[4] Li et al. (2010) http://www.cs.cornell.edu/projects/p2f/
[5] CVIT-IDH
[6] Crandall et al. (2011) http://vision.soic.indiana.edu/projects/disco/

| dataset | number of 3D points in reconstructions | | | | | | run-time for match-graph construction | | | | |
|---|---|---|---|---|---|---|---|---|---|---|---|
| | Kd-tree | | CasHash | | Our | | Kd-tree | CasHash | Our (CPU) | Our (GPU) | SIFTGPU |
| | #pts | #pts3+ | #pts | #pts3+ | #pts | #pts3+ | sec. | sec. | sec. | sec. | sec. |
| NDP | 85K | 46K | 82K | 43K | **109K** | **65K** | 6504 | 1408 | 3702 | 171 | 999 |
| TSB | 178K | 112K | 180K | 111K | **204K** | **132K** | 27511 | 8660 | 8965 | 857 | 7019 |
| BNM | 39K | 12K | 40K | 11K | **179K** | **77K** | 18282 | 3662 | 5120 | 553 | 4799 |

Table 4: Comparison of run-time for match-graph construction and number of 3D points in final models for three datasets. For NDP and TSB all images (99 and 193 resp.) are registered for all methods. For BNM, Kd-tree and CascadeHash matching based reconstructions register only 119 and 136 images respectively, while geometry-aware matching register 181 of 191 images. Also, point clouds for SFM with our method are denser.

| | | | | | repro. error | | cam. dists | |
|---|---|---|---|---|---|---|---|---|
| dataset | #cams | #pts | #pts3+ | #pairs | mean | med. | mean | med. |
| PTI | 574 | 126K | 57K | 66982 | 0.86 | 0.51 | 2.51 | 2.36 |
| PTE | 782 | 259K | 124K | 303389 | 0.76 | 0.49 | 0.811 | 0.78 |
| SPI | 953 | 301K | 140K | 227330 | 0.96 | 0.63 | 29.24 | 28.23 |
| SPE | 1155 | 380K | 180K | 575134 | 0.70 | 0.47 | 3.09 | 1.99 |
| AQD | 5147 | – | 1402K | 538131 | 0.41 | 0.30 | 179.01 | 172.77 |

Table 5: Statistics for baseline models reconstructed using Bundler with Kd-tree based pairwise matching of all features for all image pairs. '#pairs' indicate the image pairs connected by co-visible 3D points. The columns under 'cam. dists' indicate the average and median distances between the locations of the reconstructed cameras.

Table 4 compares the match-graph construction time and the number of 3D points in the final reconstruction for the three methods. Geometry-aware matching clearly outperforms other methods, as the reconstructions with our match-graphs produce denser point clouds and also recover more reliable points with track length 3 and higher. Figure 6 shows the reconstruction of BNM dataset with unguided matching and geometry-aware matching. The reconstruction with geometry- aware matching is more complete compared to other methods. The time for match-graph computation using our method is significantly less than Kd-tree based matching and only slightly worse than CascadeHash. It is worth noting that the timing given for our method is inclusive of run-time for Kd-tree based initial matching, performing initial matching using CascadeHash can reduce the overall time further. The GPU implementation of geometry-aware matching also outperforms unguided matching on GPU (Wu, 2007) giving a speedup of 5-9x.

### 7.2. Evaluation of Multistage SFM Pipeline

*Datasets and ground-truth* To evaluate our multistage SFM pipeline, we reconstruct components of ∼500-1000 images from Rome16K dataset (PTI, PTE, SPI, SPE in Table 3). For completeness of discussion, we also use our pipeline to reconstruct two large datasets of multiple structures. Hampi Vittala Temple (HVT) is a well-connected dataset comprising of ruins of multiple disjoint shrines with intricate carvings and fluted pillars spread over an area of 3 kilometers, whereas Cornell Arts Quad (AQD) is a weakly-connected dataset of multiple urban buildings. Though these datasets are reconstructed as single large components, in practice such large datasets should be divided into multiple components similar to (Frahm et al., 2010; Bhowmick et al., 2014) and our pipeline should only be used reconstruct individual components to be merged later.

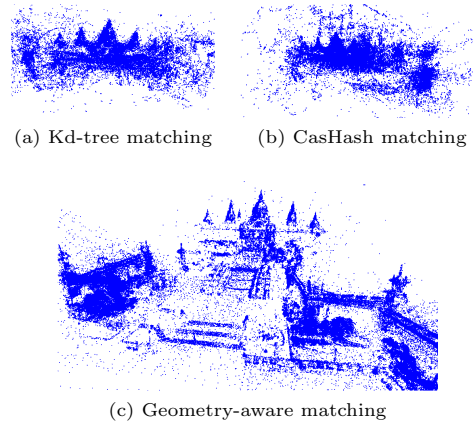

Figure 6: Reconstructions of BNM dataset using Bundler with three different match-graphs produced by, (a) Kd-tree matching, (b) CascadeHash matching, and (c) geometry-aware matching. Models with unguided matching (a,b) remain incomplete.

In absence of ground-truth, we use Bundler to reconstruct models for the Rome16K components with Kd-tree based matching of all features and consider these as the baseline models for all comparisons. For AQD dataset, we use the model reconstructed using method of Crandall et al. (2011) as baseline (provided by authors) and use the ground-truth camera positions (available for 208 cameras) for measuring absolute translation errors. Table 5 provides reconstruction statistics of the baseline models. The reconstruction of HVT dataset could not be completed using Bundler, hence we only provide qualitative comparison for this dataset.

| dataset | sfm method | #cam. | #pts. | #pts3+ | repro. error in pixels | | rot. error in degrees | | translation error abs. error | | relative error | |
|---|---|---|---|---|---|---|---|---|---|---|---|---|
| | | | | | mean | med. | mean | med. | mean | med. | mean | med. |
| PTI | BDLR | 526 | 36K | 15K | 0.97 | 0.71 | 0.133 | 0.004 | 0.239 | 0.033 | 0.095 | 0.014 |
| | VSFM | 544 | 42K | 34K | 1.80 | 1.58 | 0.014 | 0.005 | 0.120 | 0.026 | 0.048 | 0.011 |
| | V+PM | 502 | 36K | 20K | 1.74 | 1.52 | 0.015 | 0.003 | 0.125 | 0.027 | 0.049 | 0.012 |
| | GSFM* | 495 | 21K/35K | 7K/13K | 1.58 | 1.14 | 0.071 | 0.129 | 0.571 | 0.732 | 0.227 | 0.310 |
| PTE | BDLR | 780 | 52K | 27K | 0.86 | 0.62 | 0.004 | 0.002 | 0.386 | 0.003 | 0.476 | 0.004 |
| | VSFM | 782 | 49K | 33K | 1.57 | 1.42 | 0.003 | 0.001 | 0.020 | 0.003 | 0.025 | 0.004 |
| | V+PM | 782 | 49K | 33K | 1.73 | 1.44 | 0.005 | 0.001 | 0.390 | 0.106 | 0.028 | 0.005 |
| | GSFM* | 764 | 31K/52K | 14K/25K | 1.77 | 1.31 | 0.132 | 0.120 | 0.312 | 0.341 | 0.384 | 0.437 |
| SPI | BDLR | 902 | 83K | 39K | 1.26 | 0.81 | 0.299 | 0.002 | 55.258 | 0.160 | 1.889 | 0.006 |
| | VSFM | 934 | 82K | 51K | 1.66 | 1.48 | 0.005 | 0.001 | 0.390 | 0.106 | 0.013 | 0.004 |
| | V+PM | 923 | 76K | 48K | 1.86 | 1.49 | 0.006 | 0.001 | 0.821 | 0.110 | 0.028 | 0.004 |
| | GSFM* | 895 | 47K/86K | 15K/34K | 2.08 | 1.53 | 0.114 | 0.081 | 3.881 | 9.231 | 0.132 | 0.326 |
| SPE | BDLR | 1193 | 76K | 37K | 0.97 | 0.63 | 0.056 | 0.004 | 3.607 | 0.047 | 1.167 | 0.023 |
| | VSFM | 1146 | 83K | 55K | 1.34 | 1.15 | 0.006 | 0.001 | 0.292 | 0.009 | 0.095 | 0.004 |
| | V+PM | 1139 | 79K | 51K | 1.38 | 1.14 | 0.007 | 0.002 | 0.304 | 0.012 | 0.098 | 0.006 |
| | GSFM* | 1105 | 60K/96K | 22K/41K | 1.54 | 0.98 | 0.085 | 0.054 | 1.291 | 0.054 | 0.417 | 0.027 |
| HVT | V+PM | 2997 | 780K | 687K | 3.79 | 2.15 | – | – | – | – | – | – |
| AQD | V+PM | 3860 | 358K | 284K | 1.21 | 0.96 | 0.018 | 0.017 | 1.24m | 0.09m | 0.030 | 0.018 |

Table 6: Reconstruction statistics for coarse global models recovered using Bundler (BDLR), VisualSFM (VSFM) will all-pair match-graph, VisualSFM with preemptive matching based match-graph (V+PM), and Global SFM (GSFM). Most coarse models are 80%-100% complete in number of pose estimated images (#cam) but sparse in the number of 3D points compared to the baseline models. The reprojection errors are fairly low and comparable to the baseline models for all but GSFM models. The camera rotation and translation errors also indicate that coarse global models are comparable in quality to the baseline models. For AQD, the abs. translation errors are in meters w.r.t. 208 ground-truth cameras.

*Evaluation of coarse model reconstruction* For coarse model reconstruction of datasets, we use three popular SFM implementations, (i) Bundler (BDLR) – a serial implementation of the incremental SFM approach (Snavely et al., 2006) that uses ceres solver (Agarwal et al., 2010a) for its BA steps; (ii) VisualSFM (VSFM) – a highly optimized incremental SFM package that also leverages GPU for many steps, such as feature matching (Wu, 2007) and parallel bundle adjustment (Wu et al., 2011); and (iii) Theia – theia (Sweeney, 2015) is an open source SFM library with implementation of global SFM (GSFM) approach. Theia's GSFM implementation uses motion averaging for rotation estimation (Chatterjee and Govindu, 2013), state-of-the-art graph-based 1D translation estimation (Wilson and Snavely, 2014), and ceres solver (Agarwal et al., 2010a) for its BA steps.

For Bundler and Theia reconstructions, feature matching of coarse features for all image pairs is performed in parallel on a 200-core cluster. This step uses the Kd-tree based matching with hybrid selection of $\eta\%$ features as discussed in section 4. For VisualSFM reconstruction, coarse feature matching of 20% high-scale features is performed using SIFTGPU as the hybrid scheme yields little advantage for parallel distance computation based matching on GPU. We wish to show that the traditional pre-processing steps (steps 1 and/or 2 in Table 1) of large scale SFM pipeline can and should be used in connection with our method for reconstructing well-connected datasets. To demonstrate this, instead of creating a match-graph by matching coarse features of all image pairs, we create match-graphs by matching coarse features for only the image pairs selected using preemptive matching (Wu, 2013) and later perform reconstruction using VisualSFM. This method of creating the coarse models is referred by the label V+PM for brevity.

Table 6 provides the statistics of coarse models reconstructed using different methods. It can be observed that all coarse reconstruction methods are able to register between 70%-100% cameras w.r.t. the baseline models. Some GSFM coarse models are sparser with shorter tracks potentially causing insufficient camera-point visibility relationships necessary for success of the camera and point addition stages. We augment these models with more 3D points by verifying the feature matches discarded by GSFM in the early steps using the final geometry, re-triangulating the tracks formed by the added matches, and bundle adjusting the tracks. GSFM rows in Table 6 show the number of 3D points before and after augmentation. The augmented GSFM coarse models are comparable in #pts to other coarse models. For all coarse models, the mean and median reprojection errors are slightly higher than the baseline models, even for BDLR. This could be attributed to the fact that the high-scale features are localized with lesser sub-pixel accuracy than low-scale features in scale-space. The reprojection errors are still fairly low, less than 2 pixels for all models.

| dataset | sfm | #cam | #pts | #pts3+ | repro. error in pixels | | rot. error in degrees | | translation error | | | | frac. pairs |
| --- | --- | --- | --- | --- | --- | --- | --- | --- | --- | --- | --- | --- | --- |
| | | | | | | | | | abs. error | | relative error | | |
| | | | | | mean | med. | mean | med. | mean | med. | mean | med. | frac. |
| PTI | BDLR | 565 | 145K | 47K | 1.32 | 0.93 | 0.222 | 0.005 | 0.265 | 0.037 | 0.106 | 0.016 | 0.90 |
| | VSFM | 570 | 147K | 48K | 1.66 | 1.33 | 0.087 | 0.004 | 0.135 | 0.029 | 0.054 | 0.012 | 0.95 |
| | V+PM | 555 | 142K | 47K | 1.65 | 1.33 | 0.178 | 0.004 | 0.202 | 0.335 | 0.080 | 0.014 | 0.93 |
| | GSFM | 549 | 131K | 31K | 1.94 | 1.67 | 0.135 | 0.056 | 0.476 | 0.775 | 0.189 | 0.328 | 0.73 |
| PTE | BDLR | 782 | 287K | 98K | 1.15 | 0.71 | 0.005 | 0.003 | 0.385 | 0.004 | 0.475 | 0.005 | 0.99 |
| | VSFM | 782 | 279K | 96K | 1.39 | 1.02 | 0.004 | 0.001 | 0.020 | 0.004 | 0.025 | 0.005 | 0.99 |
| | V+PM | 782 | 280K | 96K | 1.36 | 1.00 | 0.004 | 0.001 | 0.023 | 0.004 | 0.028 | 0.005 | 0.99 |
| | GSFM | 782 | 296K | 82K | 1.81 | 1.52 | 0.070 | 0.048 | 0.416 | 0.445 | 0.548 | 0.570 | 0.90 |
| SPI | BDLR | 935 | 429K | 131K | 1.45 | 1.06 | 0.345 | 0.004 | 53.434 | 0.184 | 1.827 | 0.007 | 0.82 |
| | VSFM | 942 | 412K | 102K | 1.89 | 1.59 | 0.027 | 0.002 | 0.473 | 0.108 | 0.016 | 0.004 | 1.06 |
| | V+PM | 942 | 434K | 137K | 1.65 | 1.31 | 0.039 | 0.002 | 0.922 | 0.111 | 0.032 | 0.004 | 1.03 |
| | GSFM | 938 | 418K | 75K | 1.98 | 1.71 | 0.189 | 0.105 | 4.519 | 9.375 | 0.154 | 0.332 | 0.64 |
| SPE | BDLR | 1144 | 464K | 138K | 1.32 | 0.83 | 0.126 | 0.004 | 3.497 | 0.052 | 1.132 | 0.026 | 0.86 |
| | VSFM | 1154 | 468K | 150K | 1.23 | 0.83 | 0.019 | 0.001 | 0.305 | 0.009 | 0.099 | 0.005 | 0.98 |
| | V+PM | 1155 | 486K | 152K | 1.35 | 0.91 | 0.029 | 0.002 | 0.305 | 0.012 | 0.099 | 0.006 | 1.01 |
| | GSFM | 1154 | 485K | 119K | 1.81 | 1.27 | 0.129 | 0.057 | 1.496 | 0.049 | 0.484 | 0.024 | 0.81 |
| HVT | V+PM | 3003 | 4084K | 1942K | 1.74 | 1.27 | – | – | – | – | – | – | – |
| AQD | V+PM | 4429 | 1706K | 738K | 1.77 | 1.41 | 0.173 | 0.001 | 13.2m | 1.56m | 0.028 | 0.005 | 0.61 |

Table 7: Reconstruction statistics for the final models reconstructed using the multistage pipeline with coarse reconstruction using Bundler, VisualSFM, VisualSFM+PM, and Global SFM. The final models are nearly complete in number of cameras and rich in number of 3D points. The reprojection and the camera pose errors are fairly low. The last coumn gives the number of connected image pairs as a fraction of connected pairs in the baseline models, indicating comparable completeness of the final models despite matching only candidate image pairs.

We align the coarse models to the baseline models using RANSAC and measure the camera rotation and translation errors. Rotation estimation is accurate for all coarse models. The mean and median rotation errors w.r.t the baseline models are less than 0.05° for most models. The absolute translation error indicates the mean and median distances between the camera locations in coarse models and the baseline models. In absence of geo-location data, the absolute translation errors are not indicative of the true distance. Hence, we also measure relative translation errors w.r.t. the scale of the baseline model captured by the mean/median distances between locations of all cameras in that baseline model (see 'cam. dists' in Table 5). The absolute errors are divided by the camera distances in the baseline models to yield relative translation errors. For all coarse models, the mean and median relative translation errors are below 2% of the mean/med. camera distances of the baselines models.

We observe that the V+PM coarse models are comparable to the models with all pair coarse models. This observation suggests that the traditional pre-processing steps (steps 1 and/or 2 in Table 1) of a large scale SFM pipeline can and should be used before our method for reconstructing well-connected datasets.

*Evaluation of point and camera addition* We enrich and complete the coarse models using two iterations of camera and point addition stages of our multistage pipeline as explained in section 5 and section 6. The reconstruction statistics for the final models are given in Table 7.

Despite being initialized with different coarse models, all final models are nearly complete in number of cameras and have higher or comparable number of 3D points w.r.t. the baseline models, except for the AQD model. We analyze the case of AQD reconstruction later while discussing limitations of our approach. Table 8 shows the number of cameras localized in the camera addition stage and the total number of reconstructed points after the point addition stage for each iteration. The point addition stage in the first iteration considers features of all images localized in the coarse reconstruction stage as well as the first camera addition stage whereas, in the second iteration, it considers only the newly localized cameras. Due to this, the second point addition stage does not add many new 3D points, but it updates the tracks of many existing points. This ensures that the newly localized cameras form connections with the existing ones. The camera addition stage in the second iteration performs slightly worse as compared to the first iteration. We believe that this can be improved by including more distinctive points in the set cover model.

| dataset | method | Iter. 1 #cam | Iter. 1 #pts | Iter. 2 #cam | Iter. 2 #pts | #unloc. |
|---|---|---|---|---|---|---|
| PTI | BDLR | 22 | 145K | 17 | 145K | 9 |
|  | VSFM | 24 | 147K | 2 | 147K | 4 |
|  | V+PM | 45 | 142K | 8 | 142K | 19 |
|  | GSFM | 39 | 130K | 15 | 131K | 25 |
| PTE | BDLR | 2 | 287K | – | – | 0 |
|  | VSFM | 0 | 279K | – | – | 0 |
|  | V+PM | 0 | 280K | – | – | 0 |
|  | GSFM | 18 | 296K | – | – | 0 |
| SPI | BDLR | 23 | 429K | 10 | 429K | 18 |
|  | VSFM | 4 | 411K | 4 | 412K | 11 |
|  | V+PM | 12 | 434K | 7 | 434K | 11 |
|  | GSFM | 29 | 417K | 14 | 418K | 15 |
| SPE | BDLR | 42 | 464K | 9 | 464K | 10 |
|  | VSFM | 8 | 468K | – | – | 1 |
|  | V+PM | 13 | 485K | 3 | 486K | 0 |
|  | GSFM | 41 | 485K | 8 | – | 0 |
| HVT | V+PM | 6 | 4084K | – | – | – |
| AQD* | V+PM | 205 | 1668K | 249 | 1694K | 718 |

Table 8: Breakdown of added cameras and points per iteration. The columns under #cam show the number of cameras localized in camera addition stage of each iteration and the columns under #pts indicate the total number of 3D points after point addition stage of each iteration. The last column indicates the number of cameras in baseline models that remain unlocalized after two iterations of multistage pipeline. *For AQD, the third iteration localizes 115 cameras, in total adding 569 cameras to the coarse model.

After coarse model reconstruction, no additional BA steps are performed. The re-projection errors in the final models are slightly higher than the baseline models but they are comparable to or lower than the corresponding coarse models due to robust guided matching in point addition stage. The camera errors in the final models do not deviate much from that of the respective coarse models, indicating robustness of the pose estimation in camera addition stage. We observed that running bundle adjustment after our multistage pipeline reduces the overall re-projection errors slightly but does not change the camera errors significantly.

*Qualitative results* Figure 7 shows selected renders of the reconstructed components of the Rome16K dataset. The first row of images show the coarse point clouds and the images in the remaining rows show different views of the final point clouds recovered by our multistage method. Though the coarse model point clouds are very sparse, the final point clouds are rich and complete. The last two rows provide close up views of parts of the models with reference images. Figure 8 shows the selected renders of the Hampi Vitthala Temple Complex dataset (HVT). HVT dataset contains images of a large structure consisting of many sub-components. For HVT coarse reconstruction, we only use V+PM method due to practical constraints imposed by its large scale. In Figure 8, the left image in the first row shows the floor-plan of the complex structure and the right image shows the aerial view of the complete reconstruction produced by our multistage approach. Our method was able to register 3003 of 3017 images and produced a rich point cloud with roughly 4 million points. The images in the third row show that the point and camera addition stages are able to significantly enrich even very sparsely reconstructed structures in the coarse model. The images in last two rows show additional views of the two prominent structures with reference images.

*Runtime performance* After coarse model estimation, all stages of our framework are image independent and embarrassingly parallel. Such parallelism has not been a characteristic of traditional incremental SFM pipelines. Also, unlike previous methods, feature matching is divided into two stages and intertwined with localization in our reconstruction pipeline. Hence, it is not straightforward to directly equate end-to-end run-time of our pipeline with other methods. For a fair evaluation, we provide runtime for different tasks of our pipeline and the traditional pipelines under different settings. We perform our experiments on a cluster with multiple compute nodes, each node is equipped with 12 hyper-threaded Intel 2.5GHz cores. For parallel computation, we use upto 200 cores, and for sequential computation, we use a single node without multithreading. The GPU based experiments are performed on a single system with Intel core-i7 3.3GHz CPU with 6 hyper-threaded cores and Nvidia GeForce GTX 970 GPU.

Table 9 provides the run-time for feature matching under different configurations. The columns under $t_c$ indicate the time for matching using 200-core cluster, the columns under $t_g$ indicate the time for matching using a GPU. The columns in their numeric order provide feature matching run-time for the following configurations. (i) Kd-tree based matching of coarse features for all image pairs, (ii) SIFTGPU matching of coarse features for all image pairs, (iii) SIFTGPU matching of coarse features for image pairs selected using preemptive matching (PM), (iv-v) geometry-aware matching (post CGM) of all features for candidate pairs, (vi) Kd-tree based matching of all features for all pairs, (vii) SIFTGPU based matching of all features for all pairs. Geometry-aware matching is clearly faster as compared to geometry-blind unguided matching in both, cluster and GPU setup.

We compare the run-time of SFM with coarse features based match-graph and the run-time of SFM with all features match-graph using different SFM methods in Table 10. The GSFM run-time corresponds to single-threaded use, this improves 1-3x when all cores of a quad-core system are leveraged. Despite using all images, SFM with coarse match-graph is faster than SFM with full match-graph. Since, our method employs the SFM step for only coarse features based match-graph, it is clearly more advantageous.

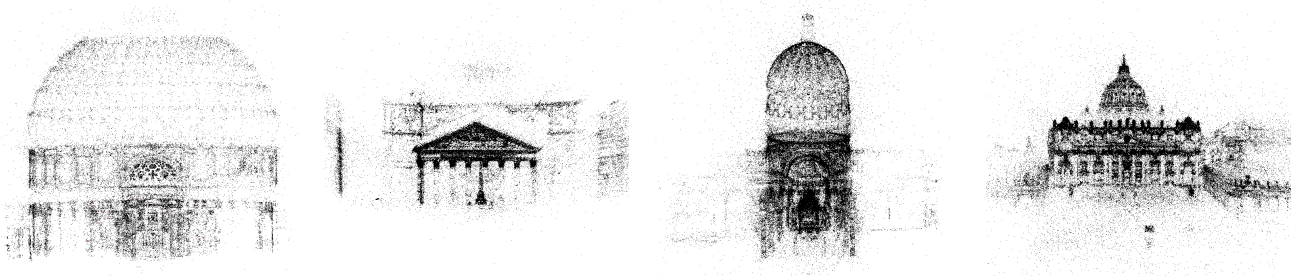

Renders of PTI, PTE, SPI, and SPE coarse reconstruction point clouds

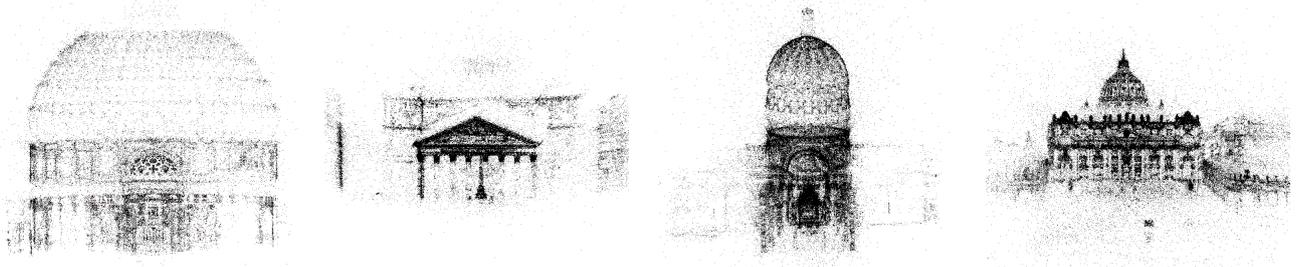

Renders of PTI, PTE, SPI, and SPE full reconstruction point clouds created by multistage pipeline

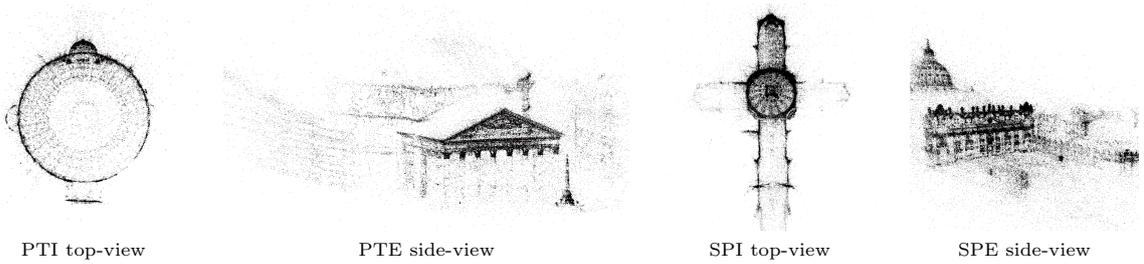

| PTI top-view | PTE side-view | SPI top-view | SPE side-view |

Additional views of PTI, PTE, SPI, and SPE full reconstruction point clouds

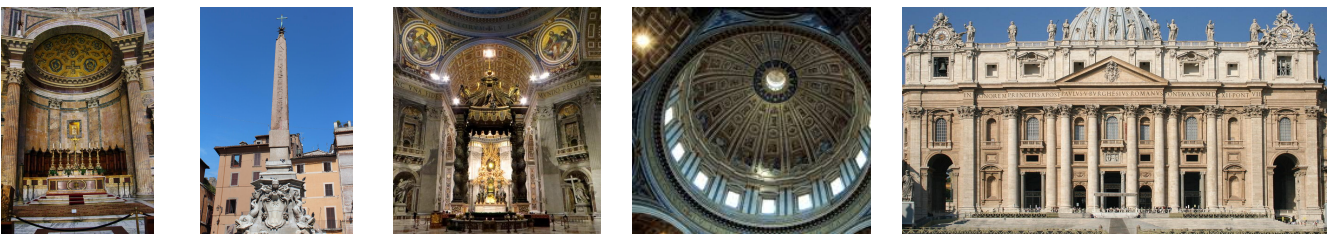

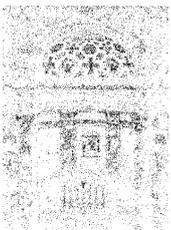 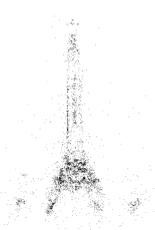 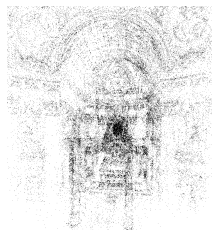 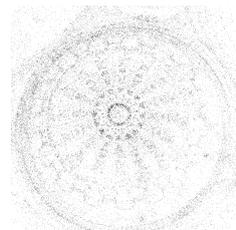 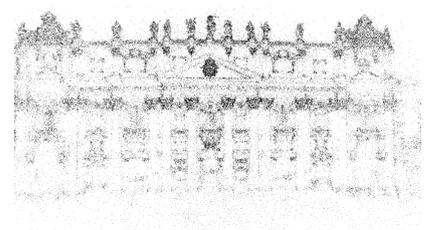

| PTI Altar | PTE Obelisk | SPI Altar | SPI Dome | SPE Facade |

Close up views of parts of the PTI, PTE, SPI, and SPE full reconstruction point clouds

Figure 7: Selected renders of reconstructed components of Rome16K dataset

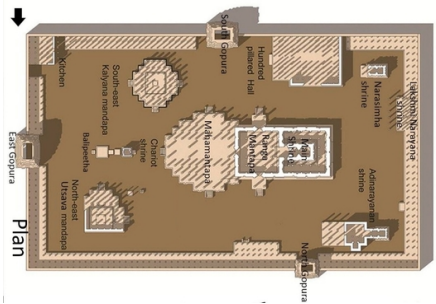
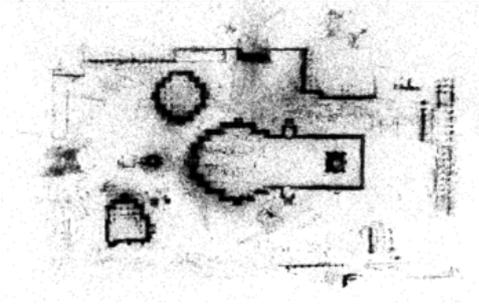

HVT floor plan  |  top-view of the full reconstruction

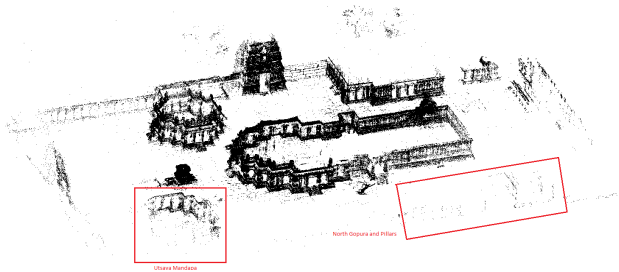
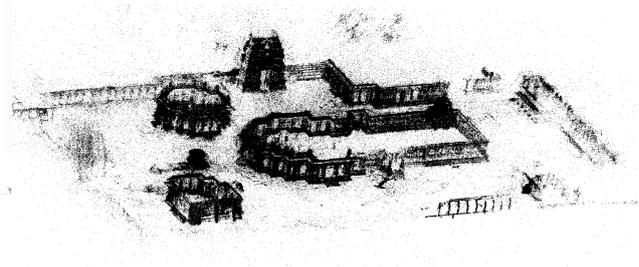

coarse reconstruction  |  full reconstruction (multistage pipeline)

Renders of Hampi Vitthala Temple Complex (HVT) point clouds. Sparsely reconstructed parts in the coarse model are highlighted by red rectangles. Close up views of these parts are given below.

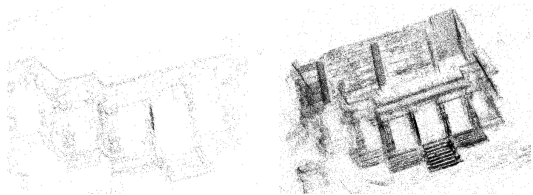
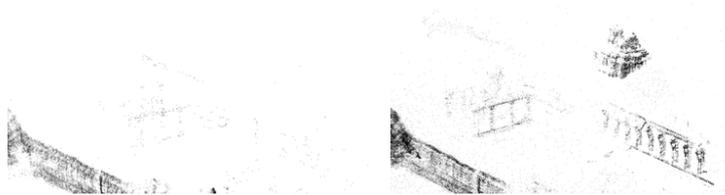

Utsava Mandapa : coarse (left), full (right)  |  North Gopura : coarse (left), full (right)

Sparsely reconstructed parts after coarse reconstruction are enriched after multistage pipeline

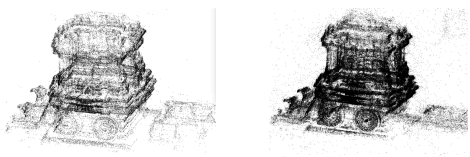
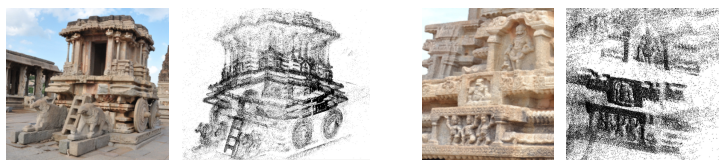

coarse (left) and full (right) reconstructions  |  Additional views of the full reconstruction with reference images

Selected renders of the stone chariot structure of HVT

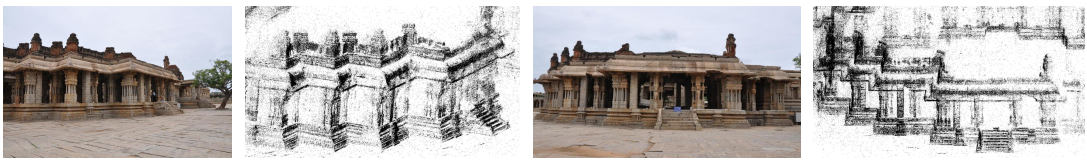

Selected renders of the Maha Mandapa (main shrine) structure of HVT

Figure 8: Point clouds of Hampi Vitthala Temple Complex reconstruction

Since, we do not perform incremental BA during later stages of the pipeline, the run-time for feature matching and coarse global model estimation dominate the total run-time of our pipeline. In comparison, the run-time for the remaining steps of our pipeline is mostly trivial.

Pre-processing steps for image localization include set cover computation, mean SIFT computation, and visual words quantization. These steps are mostly parallel and take around 10–30 seconds for Rome16K componenets on a single system with 12-cores. All steps except track merging can be performed independently for each image. Table 11 provides the run-time of these steps under maximum parallelism in number of images. At present, the sequential implementation of track merging in our pipeline takes about 100-200 seconds for the Rome16K componenets. However, in future, we wish to incorporate GPU implementation of BFS based connected-component search (Soman et al., 2010) to speed-up its performance. The end-to-end run-time of our pipeline can be computed as the total time taken by (i) 20% feature matching and geometry-aware matching of candidate pairs (Table 9), (ii) SFM for coarse reconstruction (Table 10), and (iii) the remaining steps such as 3D-2D correspondence search, localization, triangulation, etc (Table 11). Accordingly, with upto 200-core parallelism for the parallel steps and VSFM based coarse reconstruction, the total time taken by PTI and SPE reconstructions is approximately 372 and 3430 seconds.

It can be concluded that the proposed multistage arragement for SFM reconstruction can produce comparable or superior quality models as compared to the traditional methods with notable benefits in run-time. By postponing bulk of the matching until after the coarse reconstruction, we are able to leverage the known geometry for candidate image selection for matching, for efficient geometry-aware matching, and fast 3D-2D matching based localization. More importantly, the coarse reconstruction allows the rest of the steps to be performed in parallel, making it possible to speed up the reconstruction by leveraging more compute power.

*7.3. Limitations and Future Work*

The success of our framework largely depends on coverage of the coarse model. For weakly connected datasets, it is possible that the coarse reconstruction can only recover a part of the space to be modeled or results in multiple fragmented models. In this case, point and camera addition stages can only enrich the partial models and not complete it. Figure 9 shows this effect for AQD dataset. The coarse reconstruction using V+PM method results in 127 models of which 15 models have more than 25 images. The largest coarse model has 3860 images that we use as the seed model for our method (shown in Figure 9(a)). Though our method adds significant number of cameras (569) and points (∼1400K) to this seed model, the full model is missing some structures (highlighted by red rectangle in Figure 9(c)).

| dataset | 20% features | | | all features | | | |
|---|---|---|---|---|---|---|---|
| | all pairs | | PM | geom.aware* | | unguided | |
| | $t_c$ | $t_g$ | $t_g$ | $t_c$ | $t_g$ | $t_c$ | $t_g$ |
| PTI | 50 | 581 | 192 | 135 | 342 | 844 | 4135 |
| PTE | 164 | 1816 | 1280 | 778 | 1112 | 2370 | 10732 |
| SPI | 227 | 3964 | 939 | 809 | 1239 | 4129 | 16049 |
| SPE | 471 | 7920 | 3333 | 2390 | 3724 | 7591 | 28297 |
| HVT | – | – | 7207 | 1012 | – | – | 571320[7] |

Table 9: Run-time comparison for initial match-graph construction using coarse features and full match-graph construction using all features on different compute platforms. The time for geometry-aware matching of all features indicates the total time taken by matching in both iterations of the point addition stage of our pipeline. For HVT dataset, coarse feature matching without PM and unguided matching are prohibitive due to the large number of images and average number of features per image. All timings are in seconds.

| | coarse reconstruction | | | full reconstruction | | |
|---|---|---|---|---|---|---|
| dataset | BDLR | VSFM | GSFM | BDLR | VSFM | GSFM |
| PTI | 638 | 48 | 134 | 1269 | 229 | – |
| PTE | 3506 | 118 | 304 | 9043 | 646 | 832 |
| SPI | 1378 | 111 | 191 | 4481 | 206 | – |
| SPE | 4716 | 159 | 375 | 12224 | 427 | – |
| HVT | – | 1299 | – | – | 3540[7] | – |

Table 10: Run-time comparison for coarse and full reconstruction using different SFM methods, BDLR and GSFM are run on a single-core, VSFM is run on a GPU+CPU configuration. Rotation estimation with all features match-graph failed for PTI, SPI, and SPE datasets using GSFM. All timings are in seconds.

| Reconstruction Step | PTI | PTE | SPI | SPE |
|---|---|---|---|---|
| Active corres. search | 0.32 | 0.11 | 0.74 | 0.31 |
| Ranked corres. search | 0.59 | – | 1.10 | 0.76 |
| Pose estimation | 0.99 | 0.17 | 0.78 | 0.68 |
| Find candidate pairs | 2.21 | 3.67 | 4.23 | 4.98 |
| Triangulation | 0.45 | 3.17 | 1.78 | 3.07 |

Table 11: Average run-time (in sec.) taken by various image independent steps of our pipeline with maximum parallelism in num. images.

The solution is to handle weakly connected datasets as multiple overlapping components, recover separate models using our framework and combine them using shared cameras and/or points similar to the methods of step 5 in the large-scale pipeline depicted in Table 1.

Though in our experiments we found the hybrid selection of 20% coarse features sufficient for global and near-complete recovery of most structures, it can be suboptimal for very high resolution images. It remains a future work to explore methods for optimal feature selection.

In future, we would also like to explore an adaptive strategy for set cover selection that prioritizes points belonging to the newly localized cameras to further improve the effectiveness of image localization in later iterations of our pipeline.

---

[7] Reported by Bhowmick et al. (2014) on a comparable hardware

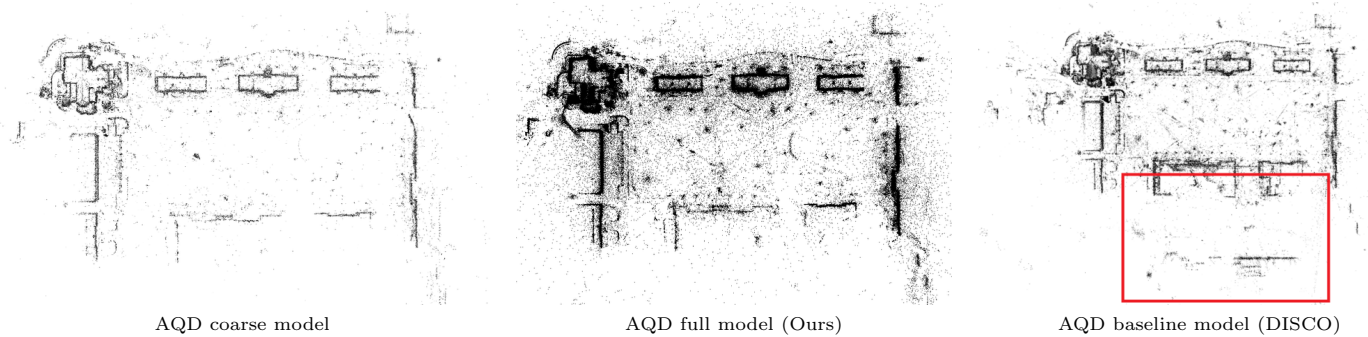

| AQD coarse model | AQD full model (Ours) | AQD baseline model (DISCO) |

Figure 9: Renders of Cornell Arts Quad reconstruction : (a) shows the largest recovered model using V+PM method and 20% features ($|C| = 3860$, $|P| = 3528$); (b) shows the result of our multistage pipeline ($|C| = 4429$, $|P| = 1706K$); (c) shows the baseline model reconstructed by Crandall et al. (2011) ($|C| = 5147$, $|P| = 1407K$). The parts of the baseline model highlighted by red rectangle could not be recovered using our method.

## 8. Conclusion

In this paper, we presented a multistage approach as an alternative to match-graph construction and structure from motion steps of the traditional large scale 3D reconstruction pipeline. The proposed approach provides an opportunity to leverage the constraints captured by the coarse geometry for making the rest of the processing more efficient and parallel. We evaluated our method comprehensively and showed that it can produce similar or better quality reconstructions as compared to the traditional methods while being notably fast. In future, we wish to explore real-time reconstruction applications with multistaging and also the possibility of extending our framework for performing fast multi-view stereo.

## Acknowledgements

This work is supported by Google India PhD Fellowship and India Digital Heritage Project of the Department of Science and Technology, India. We would like to thank Vanshika Srivastava for her contributions to the project and Chris Sweeney for his crucial help regarding use of Theia for our experiments. We would also like to thank the authors of Bhowmick et al. (2014) for sharing the curated image set of Hampi Vitthala Temple.

## Appendix A. GPU implementation of geometry-aware feature matching

The algorithm for geometry-aware matching described in subsection 6.2 is well suited for parallel computation on the GPU. We use high performance CUDA parallel primitives like sort and reduce from the Thrust (Hoberock and Bell (2010)) library and obtain a significant speedup over CPU. Exploiting parallelism in Kd-tree based hierarchical search on high-dimensional data is difficult. GPU flann (Muja and Lowe (2014)) works only for up to 3-dim data. The guided-search steps are made parallel by leveraging point/group independence as follows.

*Step 1. Grid computation and feature binning* We launch a grid of threads on the GPU with one thread processing one feature each. Each thread computes the cell centers and the cell indices for all features in target image in parallel. A fast parallel sort using cell indices as keys brings all features belonging to the same cell together. Using a fast parallel prefix scan yields the starting index into each cell of the sorted array.

*Step 2. Epipolar line based feature clustering* In this step also a grid of threads is launched where each thread computes the epipolar line and its intersection with the image bounding box for each feature in source image. To cluster the features in the source image based on their epipolar lines, we perform a parallel sort of all the features using the coordinates of the line-image intersection points as the keys. To assign sorted feature points to clusters, we again launch a grid of threads where each thread writes 0 or 1 to a stencil array if its epipolar line differs from the previous one by more than 2 pixels. We then perform a fast parallel scan on this stencil array to provide us with the individual cluster indices.

*Step 3. Finding the set of candidate matches* We use one CUDA block per cluster of features in source image to find its corresponding set of candidate matches in the target image. Each thread in the block takes one equidistant point on the epipolar line, computes its corresponding cell indices and retrieves the features binned into these cells in step 1. The candidate matches corresponding to each cluster of query features are stored in the GPU global memory.

*Step 4. Finding the true match from candidates* For every query point, we need to find its two nearest points in the respective candidates set derived in step 3. Each query point is handled by one CUDA thread block and each thread within the block computes the $L_2$ distance between the query feature and one feature from the candidate set in parallel. A parallel block-wise minimum and next-minimum is computed and followed by a ratio test to declare a match.